\documentclass[11pt,a4paper]{article}
\pdfoutput=1
\usepackage[hyperref]{acl2017}
\usepackage{times}
\usepackage{latexsym}

\usepackage{url}

\usepackage{amsmath}

\usepackage{todonotes}

\usepackage[utf8]{inputenc}

\usepackage{booktabs}

\aclfinalcopy % Uncomment this line for the final submission
\newcommand{\smallcapsfont}[1]{\begin{small}\textsc{\texttt{#1}}\end{small}}

\DeclareMathOperator*{\argmax}{arg\,max}

\title{Morphological Embeddings for Named Entity Recognition in Morphologically Rich Languages}

\author{Onur Güngör \\
  Boğaziçi University, Istanbul \\
  Huawei R\&D Center, Istanbul \\
  {\tt onurgu@boun.edu.tr} \\[3ex]
  Suzan Üsküdarlı \\
  Boğaziçi University, Istanbul \\
  {\tt suzan.uskudarli@boun.edu.tr} \\
  \\\And
  Eray Yıldız \\
  Huawei R\&D Center, Istanbul \\
  {\tt eray.yildiz@huawei.com} \\
  \\[3ex]
  Tunga Güngör \\
  Boğaziçi University, Istanbul \\
  {\tt gungort@boun.edu.tr} \\}

\date{}

\begin{document}
\maketitle

\begin{abstract}
In this work, we present new state-of-the-art results of 93.59\% and 79.59\% for Turkish and Czech named entity recognition based on the model of \cite{lample2016neural}. We contribute by 
proposing several schemes for representing the morphological analysis of a word in the context of named entity recognition.
We show that a concatenation of this representation with the word and character embeddings improves the performance.
The effect of these representation schemes on the tagging performance is also investigated.
  
\end{abstract}

\section{Introduction}
\label{sec:introduction}

Named Entity Recognition (NER) is an important task in Natural Language Processing (NLP) that aims to discover references to entities in some text. Identified entities are classified into predefined categories like person, location and organization. NER is mostly utilized prior to complex natural language understanding tasks such as relation extraction, knowledge base generation, and question answering~\cite{Liu_2011,Lee_2007}. Additionally, NER systems are often part of search engines~\cite{guo2009named} and machine translation systems~\cite{babych2003improving}.

Early studies regarding NER propose using hand crafted rules and lists of names of people, places and organizations~\cite{humphreys1998university,appelt1995sri}. Traditional approaches typically use several hand-crafted features such as capitalization, word length, gazetteer related features, and syntactic features (part-of-speech tags, chunk tags, etc.) \cite{mccallum2003early,finkel2005incorporating}. A wide range of machine learning-based methods have been proposed to address named entity recognition. Some of the well known approaches are conditional random fields (CRF)~\cite {mccallum2003early,finkel2005incorporating}, maximum entropy~\cite{borthwick1999maximum}, bootstrapping~\cite{jiang2007instance,wu2009domain}, latent semantic association~\cite{guo2009domain}, and decision trees~\cite{szarvas2006multilingual}.

Recently, deep learning models have been instrumental in deciding how the parts of the input should be composed to allow the most beneficial features to form leading to state-of-the-art results \cite{collobert2011natural}.
Likewise, researchers have found that representing words with fixed length vectors in a dense space helps improving the overall performance of many tasks: sentiment analysis~\cite{socher2013recursive}, syntactic parsing~\cite{Collobert_2008}, language modeling~\cite{mikolov2010recurrent}, part-of-speech tagging and NER~\cite{collobert2011natural}. These word representations or embeddings are automatically learned both during or before the training using various methods such as Word2Vec~\cite{mikolov2013distributed} and GloVe~\cite{pennington2014glove}.

Building upon these findings, there are recent studies which treat the NER task as a sequence labeling problem which employ LSTM or GRU components \cite{lample2016neural,huang2015bidirectional,ma2016end,Yang2016MultiTaskCS} to capture the syntactic and semantic relations between the units that make up a natural language sentence. However, these approaches are not well studied for morphologically rich languages. Unlike other languages, morphologically rich languages such as Turkish may retain important information in the morphology of the surface form of the word while the same information may be contained in the syntax of other languages. For example, the word ``İstanbul'daydı'' means `he/she was in Istanbul' in English. The morphological analysis of the word is ``İstanbul+\allowbreak Noun+\allowbreak Prop+\allowbreak A3sg+\allowbreak Pnon+\allowbreak Loc\string^DB+\allowbreak Verb+\allowbreak Zero+\allowbreak Past+\allowbreak A3sg'' where `Prop' indicates a proper noun, `A3sg' signifies the third singular person aggreement whereas `Pnon' signifies no possesive agreement is active. `DB' indicates a transition of Part-Of-Speech type usually induced by a derivative suffix \cite{oflazer1994two}. In this case, the derivation is triggered by the `-di' suffix which was decoded as `Past' tag which indicates past tense. 
As seen from the example, morphological tags may help in capturing syntactic and semantic information. In order to address this, character based embeddings in word representations \cite{lample2016neural} and entities tagged at character level \cite{kuru-can-yuret:2016:COLING} were proposed for NER. 
Embedding based frameworks for representing morphology were also proposed in other contexts such as language modeling \cite{luong2013better,Santos2014LearningCR,Xu2017ImplicitlyIM,Bhatia2016MorphologicalPF,Lankinen2016ACC} and morphological tagging and segmentation \cite{theroleofcontextshen-EtAl:2016:COLING1,Cotterell2017JointSS}.
However, even though morphological tags have been employed in the past \cite{tur2003statistical,Yeniterzi:2011:EMT:2000976.2000995}, our work is the first to propose an embedding based framework for representing the morphological analysis in the context of NER. 

We build upon a state-of-the-art NER tagger \cite{lample2016neural} based on a sequential neural model with extensible word representations in Section \ref{sec:model}.
We show that augmenting the word representation with morphological embeddings based on Bi-LSTMs (Section \ref{sec:embeddings}) improves the performance of the base model, which uses only pretrained word embeddings.
We contribute by investigating various configurations of the morphological analysis of the surface form of a word (Section \ref{sec:configurations}).
In Section \ref{sec:results}, we compare the performance of several experiment setups which employ character and morphological embeddings in various combinations. We report F1-measures of 93.59\% and 79.59\% for Turkish and Czech respectively. These results are the state-of-the-art results compared to previous work \cite{demir2014improving,seker-eryigit:2012:PAPERS} which rely on a regularized averaged perceptron and CRF respectively both with hand crafted features.

\section{Model}
\label{sec:model}

We formally define an input sentence as $X = (x_1, x_2, \dots, x_n)$ where each $x_i$ is a fixed length vector of size $d$, consisting of embeddings that represent the $i$th word (See Section~\ref{sec:embeddings}). $x_i$ are then fed to a Bi-LSTM which is composed of two LSTMs \cite{Hochreiter1997} treating the input forwards and backwards respectively. Thus we obtain these forward and backward components' cell matrices $\overrightarrow{H}$ and $\overleftarrow{H}$ which are both of size $n \times p$, where $p$ is the number of dimensions of one component of the Bi-LSTM. Thus, $\overrightarrow{H}_{i,j}$ is the value of $j$th dimension of $i$th output vector of the right component which corresponds to the $i$th word in the sentence.
We then feed the concatenation of these matrices $H = [ \overrightarrow{H}, \overleftarrow{H}]$ to a fully connected hidden layer of $K$ output neurons.

To model the dependencies between the corresponding labels of consecutive input units,
we follow a conditional random field (CRF) \cite{lafferty2001conditional} based approach. This dependency is clearly indicated by labels in IOB tagging scheme, i.e. B-PERSON, I-PERSON, etc.

To do this, we
obtain a score vector at each position $i$ and aim to minimize the following objective function for a single sample sentence $X$:
\[
    s(X, y) = \sum_{i} A_{y_i,y_{i+1}} + \sum_i \xi_{i,y_i}
\]
where $A_{i,j}$ represents the score of a transition from tag $i$ to $j$ and $\xi_i$ are the tag scores at position $i$ output by the uppermost fully connected layer. Using this model, we decode the most probable tagging sequence $y^*$ as $\argmax_{\tilde{y}} s(X, \tilde{y})$.

\subsection{Embeddings}
\label{sec:embeddings}

\begin{figure}[h]
  \centering
  \includegraphics[width=0.4\textwidth]{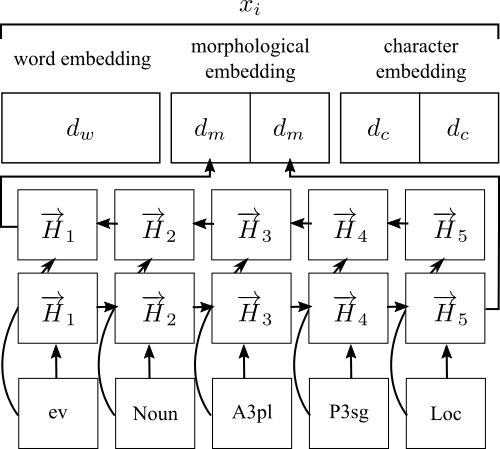}
  \caption{Bi-LSTM component for morphological embeddings.}\label{fig:embeddings}
\end{figure}

It has been shown that modeling units of information in a natural language input as fixed length vectors is more effective at encoding semantic properties of the words compared to deciding on the features apriori~\cite{collobert2011natural,Turian2010WordRA}.
Therefore we represent the input words, $x_i$, as a combination of three embeddings: \textit{word}, \textit{character}, and \textit{morphological}. Thus the size $d$ of $x_i$ is $d_w + 2d_m + 2d_c$. We describe these embeddings below and illustrate in Figure~\ref{fig:embeddings}.

\textbf{Word embeddings.} A vector of size $d_w$ which is learned by the global objective function. However, we never learn this component from scratch, instead we load pretrained vectors.

\textbf{Character embeddings.} We learn another fixed length vector of size $2d_c$ for each word. However, in contrast with a word embedding, we want to capture the covert relationships in the sequence of characters of the word. To achieve this, we have a separate Bi-LSTM component for this embedding type with a cell dimension of $d_c$. We feed it with the characters of the surface form of $x_i$ and concatenate the cell output of the forward and backward LSTMs to obtain the \textit{character} embedding of the word.

\textbf{Morphological embeddings.} These are constructed similar to \textit{character} embeddings. In this case, the individual tags of the morphological analysis are treated as a sequence and fed into the separate Bi-LSTM component for \textit{morphological} embeddings to obtain a vector of length $2d_m$. We devised several different combinations of morphological tags which is explained in Section \ref{sec:configurations}. To illustrate, we use the word `evlerinde' which can both mean `in their house' or `in their houses' or `in his/her houses' in Turkish. In Figure \ref{fig:embeddings}, we assume that the correct morphological analysis is `ev+Noun+A3pl+P3sg+Loc', where `A3pl' indicates $3^{\text{rd}}$ person plural, `P3sg' is the possessive marker for $3^{\text{rd}}$ person singular, and `Loc' is the locative case marker, thus can be translated as `in his/her houses'.

\subsection{Embedding configurations}
\label{sec:configurations}

We experimented with different combinations of morphological tags to discover an effective configuration for extracting the syntactic and semantic information in the morphological analysis of a token.

A simple embedding configuration is to use all morphological tags in the analysis along with or without the root (we call this \smallcapsfont{WR} and \smallcapsfont{WOR} respectively). Secondly we tried to remove the tags between the root and the derivation boundary (DB) based on the information they carry may not be relevant in some aspects because of the transformation into a new word with partly different lexical and syntactic properties. We call this version \smallcapsfont{WR\_ADB}, i.e. ``İstanbul+\string^DB+Verb+\allowbreak Zero+\allowbreak Past+\allowbreak A3sg''. Lastly, we devised a scheme in which we treat the string of morphological analysis as a surface form and process each character of this surface form as we did for \textit{character} embeddings and call this scheme as \smallcapsfont{char}.

\section{Experiments}
\label{sec:experiments}

\subsection{Training}

The parameters to be learned by the training algorithm is the parameters of the Bi-LSTM in Section \ref{sec:model}, the parameters of the Bi-LSTMs for the \textit{character} and \textit{morphological} embeddings and the word embeddings for each unique word. We experimented with several different choices for the number of dimensions for these parameters and observed that a choice of 100 for word embeddings, 200 for character embeddings and 200 for morphological embeddings. We trained the models by calculating the gradients using backpropagation algorithm and updating with stochastic gradient descent algorithm with a learning rate of 0.01. We also employed gradient clipping to handle gradients diverging from zero. We additionally used dropout on the inputs with probability 0.5.

\subsection{Dataset}
\label{sec:dataset}

We train and evaluate our model with a corpus which is widely used in previous work on Turkish NER ~\cite{tur2003statistical}. 
In addition to the entity tags and tokens, the corpus also contains the disambiguated morphological tags of input words. We observed many morphological analysis errors and incorrect entity taggings in the corpus \cite{tur2003statistical}, probably as a result of automated analysis and labeling.

We obtained word embeddings\footnote{We will make these word embeddings available at (we refrain from sharing the url during the review process).} of Turkish words as vectors of length 100 using the skipgram algorithm~\cite{mikolov2013distributed} on a corpus of 951M words~\cite{erayyildiz2016morphology}, 2,045,040 of which  are unique. This corpus consists of Turkish text extracted from several national newspapers, news sites, and book transcripts.

\begin{table}[t]
\centering
\begin{tabular}{|c|c|l|c|}
 \multicolumn{4}{c}{\textbf{Turkish}}  \\
\hline
 & \multicolumn{2}{c|}{\textbf{WE (pretrained)}} &                \\
   \hline
\textbf{Setup No}   & \textbf{CE}    & \textbf{ME} & \textbf{F1}\\
   \hline
1  & -    & \multicolumn{1}{c|}{-} & 90.96        \\
2  & -    & ME(\smallcapsfont{WR\_ADB})   & 90.76 \\
3  & -    & ME(\smallcapsfont{WR})              & 90.33 \\
4  & -    & ME(\smallcapsfont{char})                    & 92.79 \\
5  & -    & ME(\smallcapsfont{WOR})                & 91.18 \\
6  & CE    & ME(\smallcapsfont{WR\_ADB})   & 91.37 \\
7  & CE    & ME(\smallcapsfont{WR})              & 91.09 \\
8  & CE    & ME(\smallcapsfont{char})                    & 92.93 \\
9  & CE    & ME(\smallcapsfont{WOR})                & \textbf{93.59} \\
10 & CE    & \multicolumn{1}{c|}{-}                          & 93.37 \\
\hline
 \multicolumn{4}{c}{\textbf{Czech}}  \\
 \hline
11 & CE    & ME(\smallcapsfont{char})                          & \textbf{79.59} \\
\hline
\end{tabular}
\caption{Best performing experiment setups.}\label{tab:modelconfs}
\end{table}

\subsection{Results}
\label{sec:results}

We observed that using pretrained word embeddings gave the best results compared to learning word embeddings while training the model. Therefore we only include the results of experiment setups with pretrained word embeddings in Table \ref{tab:modelconfs}.

We start with comparing Setup 1 and 4 (and 5) and suggest that the morphological analysis does indeed contribute to higher performance with \smallcapsfont{char} and \smallcapsfont{WOR}. However, Setup 2 and 3 did not reach the performance level of Setup 1. We suspect that the reason is the relatively high number of parameters when we include the 20030 roots into the model. This effect is also seen in Setup 6 and 7 which have a lower performance compared to Setups 8 and 9.
However, we have to note that using character embeddings alone also improved the performance in Setup 10. Nevertheless, we see that the best performance is achieved in Setup 9 when both of them are employed. 
However, when we performed the McNemar's test~\cite{dietterich1998approximate}, we observed that the difference between them is not significant at 95\% confidence level.
We explain the difference in performance between Setup 4 and 5 with the errors in the morphological analysis which are mostly due to unknown or mispelled words. In those cases, the analysis become usually the same nominal case with 3rd person singular. We suspect that the fact that ME(\smallcapsfont{CHAR}) can process the root even if it is faulty allows it to capture useful information into the embedding.

Despite CE caused a large improvement generally, it provides a relatively small increase in Setup 8 compared to Setup 4. We believe that the reason behind this is that CE and ME(\smallcapsfont{CHAR}) competes with each other in representing the morphological information of the word.
The reason that Setup 9 achieved higher performance compared to Setup 8 is probably because the missing roots in Setup 9 can be covered by CE combined with relatively lower complexity of ME(\smallcapsfont{WOR}).

We have also evaluated our model on text in Czech which is another morphologically rich language. To be able to compare our results, we used the CNEC 2.0 corpus in CoNLL format as other studies did \cite{Konkol:2013}. We chose to include only the ME(\smallcapsfont{CHAR}) setup for Czech because it gave good results both with and without character embeddings.

Lastly, we compare our best results with previous state-of-the-art in Table \ref{tab:otherwork}. The performance of \cite{seker-eryigit:2012:PAPERS} without gazetteers is 89.55\%, \cite{kuru-can-yuret:2016:COLING} does not employ any external data and \cite{demir2014improving} still relies on hand-crafted features despite exploiting word embeddings trained externally.
\begin{table}[h]
\centering
\begin{tabular}{|p{4cm}|c|c|}
\hline
 & \multicolumn{2}{c|}{\textbf{F1-Measure}}  \\
\hline
\textbf{Model}                  & \textbf{Turkish} & \textbf{Czech} \\
\hline
\cite{kuru-can-yuret:2016:COLING}      & 91.30 & 72.19 \\
\hline
\cite{demir2014improving} & 91.85 & 75.61 \\
\hline
\cite{seker-eryigit:2012:PAPERS} & 91.94 & N/A \\
\hline
This work & \textbf{93.59} & \textbf{79.59} \\
\hline
\end{tabular}
\caption{Comparison with previous work.}\label{tab:otherwork}
\end{table}

\section{Conclusions}

In this work, we demonstrated a new state-of-the-art system for Turkish and Czech named entity recognition using the model of ~\cite{lample2016neural}. We introduced embedding configurations to understand the affect of different combinations of the morphological tags. Using these configurations, we showed that augmenting word representations with morphological embeddings improves the performance. However, the contribution of morphological embeddings seems to be subsumed by character embeddings in some of these configurations. Thus a thorough examination and comparison of character and morphological embeddings learned in this sense is required for further discussion.

\bibliography{acl2017}
\bibliographystyle{acl_natbib}

\end{document}